\documentclass[journal]{IEEEtran}

\usepackage[numbers,sort&compress]{natbib} 
\usepackage{float}

\usepackage{multirow}
\usepackage{threeparttable}

\usepackage[ruled,linesnumbered]{algorithm2e} 
\usepackage{amsmath}
\usepackage{amsfonts}
\usepackage{amssymb}  

\usepackage{graphicx}
\graphicspath{{images/}}                     
\usepackage{makecell}
\usepackage{hyperref}

\usepackage[utf8]{inputenc}

\usepackage{pifont}

\usepackage{booktabs}

\usepackage{soul}
\usepackage{color, xcolor}
\soulregister\cite7
\soulregister\citep7
\soulregister\citet7 
\soulregister\ref7 
\soulregister\pageref7 

\begin{document}
%


\title{CDMamba: Remote Sensing Image Change Detection with Mamba}

\author{Haotian Zhang$^{1}$, Keyan Chen$^{1}$, Chenyang Liu$^{1}$, Hao Chen$^{2}$, Zhengxia Zou$^{1}$, and Zhenwei Shi$^{1, \star}$ 
\\
\vspace{6pt}
Beihang University$^1$, Shanghai Artificial Intelligence Laboratory$^2$
}
\date{April. 2024}

\maketitle
\begin{abstract}

Recently, the Mamba architecture based on state space models has demonstrated remarkable performance in a series of natural language processing tasks and has been rapidly applied to remote sensing change detection (CD) tasks. However, most methods enhance the global receptive field by directly modifying the scanning mode of Mamba, neglecting the crucial role that local information plays in dense prediction tasks (e.g., CD). In this article, we propose a model called CDMamba, which effectively combines global and local features for handling CD tasks. Specifically, the Scaled Residual ConvMamba (SRCM) block is proposed to utilize the ability of Mamba to extract global features and convolution to enhance the local details, to alleviate the issue that current Mamba-based methods lack detailed clues and are difficult to achieve fine detection in dense prediction tasks. Furthermore, considering the characteristics of bi-temporal feature interaction required for CD, the Adaptive Global Local Guided Fusion (AGLGF) block is proposed to dynamically facilitate the bi-temporal interaction guided by other temporal global/local features. Our intuition is that more discriminative change features can be acquired with the guidance of other temporal features. Extensive experiments on three datasets demonstrate that our proposed CDMamba outperforms the current state-of-the-art methods. Our code will be open-sourced at https://github.com/zmoka-zht/CDMamba.

\end{abstract}


\IEEEpeerreviewmaketitle


\section{Introduction}
\label{sec:intro}

\IEEEPARstart{C}{hange} detection has become a popular research field in the remote sensing community due to the continuous development of remote sensing technology. The objective of this task is to monitor surface changes in the same area employing remote sensing images acquired at different times. Change detection plays an essential role in various fields such as urban planning \cite{Chen2020STA, coppin2002digital, wellmann2020remote, zhang2024bifa}, land cover analysis \cite{Li2016SLC-CD}, disaster assessment \cite{zheng2021building, xu2019building}, ecosystem monitoring \cite{todd1977urban, singh1986change, jackson1983spectral, chen2021building}, and resource management \cite{asadzadeh2022uav}.

Optical high-resolution remote sensing images are widely used in the field of change detection due to their ability to provide abundant detailed features such as textural and geometric structural information. However, the improvement in spatial resolution of remote-sensing images has increased the heterogeneity of the same region, which greatly limits the effectiveness of traditional change detection methods that rely heavily on empirically designed approaches (such as algebra-based \cite{jackson1983spectral, singh1986change}, transformation-based \cite{celik2009pca, han2007tasselledcap, saha2019CVA}, and classification-based methods \cite{sun2021KNN, negri2020SVM}) in dealing with complex ground conditions.

The development of deep learning technology has brought a promising new solution to the field of change detection, significantly boosting both detection accuracy and efficiency. Since Daudt et al. \cite{daudt2018FC-EF} introduced the fully convolutional network (FCN) into the field of change detection, CNN-based change detection networks have dominated for a period of time. Several representative works have been proposed \cite{zhang2020DSIFN, peng2019UNet++, fang2021snunet} by combining the characteristics of change detection tasks. For example, Zhang et al. proposed DSIFN \cite{zhang2020DSIFN}, a change detection network built using CNN combined with deep supervision. Fang et al. proposed SNUNET \cite{fang2021snunet}, employing dense connections to learn the spatiotemporal relationships of deep features. Despite the aforementioned methods achieving satisfactory results, the inherent limitations of CNN structure (insufficient global modeling ability due to receptive field restrictions) make it challenging to achieve accurate recognition in complex scenes with varying spatial and temporal resolutions. 

The rapid development of visual Transformers \cite{vit, CAT, swintransformer, zhang2022context} has provided a solution to the aforementioned issues. Specifically, by leveraging the self-attention mechanism in the vision Transformer, it effectively models the relationship between any area and the entire image, therefore addressing the issue of insufficient receptive fields in CNNs. Nowadays, an increasing number of methods are incorporating the Transformer model into change detection tasks \cite{zhang2022swinsunet, changeformer, BIT, MSCANet, DMINet, ICIFNet}. For example, Chen et al. \cite{BIT} utilized a Transformer to construct a Bi-temporal Image Transformer module to capture global spatiotemporal relationships. Bandara et al. \cite{changeformer} proposed Changeformer, which utilizes a variant of the vision Transformer to construct a backbone for extracting bi-temporal image features. Another similar work is Swinsunet proposed by Zhang et al. \cite{zhang2022swinsunet}.

However, the complexity of using Transformers for image processing scales quadratically with the length of image patches. This results in significant computational cost, making it unfriendly for tasks such as dense prediction tasks like change detection. Some methods aim to improve computational efficiency by limiting the window size \cite{swintransformer, swintransformerv2, cswin} or utilizing sparse attention mechanisms \cite{PVT, segformer, metaformer, child2019generating, random_transformer}. However, these approaches come at the cost of imposing limitations on the global receptive field. Recently, Mamba \cite{mamba}, which introduces time-varying parameters into State Space Models (SSMs) enabling data-dependent global modeling with linear complexity, has achieved significant success in the field of natural language processing and is considered an effective alternative to Transformer. Inspired by the success, the Mamba architecture has been expanded into the field of computer vision and has shown promising results in some visual tasks \cite{visionmamba, vmamba, LightM-UNet, MIM-ISTD, RSCaMa, plainmamba, Vm-unet, vivim}. Most of these methods directly modify the scanning model of the Mamba to enhance the global receptive field and capture more comprehensive global features of images. However, in dense prediction tasks such as change detection, local information plays an essential part in accurate detection. Developing an effective structure based on Mamba that integrates \textbf{global} and \textbf{local} information is valuable for advancing research in the change detection community.

In the paper, we proposed Change Detection Mamba (CDMamba), a simple yet effective model that combines global and local features to handle change detection tasks. Specifically, CDMamba is mainly composed of Scaled Residual ConvMamba (SRCM) and Adaptive Global Local Guided Fusion (AGLGF) block. Different from current methods relying solely on vanilla Mamba, the SRCM incorporates the design of locality and is designed to effectively extract global and local clues from images, aiming to alleviate the challenge of existing Mamba-based methods lacking detailed features for achieving fine-grained detection. Furthermore, considering the requirement for interaction between bi-temporal features in change detection tasks, AGLGF is designed to facilitate global/local feature-guided bi-temporal interaction. By guiding from another temporal image, the model is prompted to focus more on the change region, thereby further acquiring discriminative differential features.

In summary, the main contributions of this article are as follows:
\begin{itemize}
\item Proposed a novel CD network CDMamba, which effectively integrates global and local information utilizing the Scaled Residual ConvMamba (SRCM) module and alleviates the challenge of lacking local clues in Mamba when handling dense predict tasks (e.g., CD tasks).

\item Proposed an Adaptive Global Local Guided Fusion (AGLGF) block, which dynamically integrates global/local feature fusion guided by another temporal image to extract more discriminative change features for CD tasks.

\item Qualitative and quantitative studies on three datasets, WHU-CD, LEVIR-CD, and LEVIR+CD, show that our proposed CDMamba achieves state-of-the-art results.
\end{itemize}

The rest of this paper is organized as follows. Section \ref{sec:relatedwork} describes the related work. Section \ref{sec:method} gives the details of our proposed method. Some experimental results are reported in section \ref{sec:experiment}. And the conclusion is made in Section \ref{sec:conclusion}.


\section{Related Work} \label{relatedwork}
\label{sec:relatedwork}

\subsection{CNN-based CD Models}

With the flourishing development of deep learning technology, CNNs have gained common attention because of their excellent capability in extracting local features and have been applied in the early stages of the CD field. Daudt et al. \cite{daudt2018FC-EF}, as a pioneer, introduced the approach of FCN to propose FC-Siam-Conc, which concatenates bi-temporal images along the channel dimension and processes them as a single input, as well as two variants, FC-EF and FC-Siam-Diff, which utilize a Siamese CNN to handle bi-temporal inputs. Fang et al. \cite{fang2021snunet} proposed a densely connected CNN network for comprehensive interaction of bi-temporal image features. Zhang et al. \cite{zhang2020DSIFN} achieved multi-level fine-grained detection by applying supervision (e.g., deep supervised) to the differential features extracted by different stages of CNN. Shi et al. \cite{shi2021deeply} improved the method based on deep supervision by incorporating an attention mechanism module to acquire more discriminative features. Lei et al. \cite{lei2021difference} proposed a differential enhancement network to effectively learn the difference representation between foreground and background, in order to reduce the impact of irrelevant factors on the detection results. To learn more discriminative object-level features, Liu et al. \cite{liu2020building} proposed a dual-task constrained deep siamese convolutional network to achieve this objective. In pursuit of the same objective, Liu et al. \cite{liu2021super} proposed a super-resolution-based change detection model to alleviate the cumulative errors in bi-temporal images of different resolutions employing adversarial learning. Jiang et al. \cite{jiang2022wricnet} proposed a weighted multiscale encoding network that accurately detects different scales of change regions (e.g., large change regions or small change regions) by adaptively weighting multiscale features. Concurrently, Huang et al. \cite{huang2021multiple} achieved selective fusion of multi-temporal features by constructing MASNet based on selective convolutional kernels and multiple attention mechanisms. Zhang et al. \cite{zhang2021escnet} proposed a method that combines superpixel sampling network with CNN to reduce potential noise in pixel-level feature maps. Lv et al. \cite{lv2022spatial} employed an adaptively generated change magnitude image (CMI) to guide the learning of the change detection model, aiming to preserve the shape and size of the changing regions. 

However, despite the effectiveness of the aforementioned methods, the inherent local receptive field attributes of CNNs make it difficult to capture long-range dependencies. This limitation is fundamental in CD tasks where the changing objects are sparse. In this article, we combine the recently proposed Mamba \cite{mamba}, which has excellent long-distance modeling capabilities, to construct a change detection network to alleviate the aforementioned issue.

\subsection{Transformer-based CD Models}

With the rise of Transformers \cite{vit, chen2024rsprompter} in computer vision tasks, their beneficial capability in modeling long-range dependencies has drawn attention in the field of CD. Chen et al. \cite{BIT} proposed the BIT (Bi-temporal Image Transformer) model, which introduces Transformers into the field of change detection. It achieves efficient context modeling by sparsifying bi-temporal features into visual tokens. Zhang et al. \cite{zhang2022swinsunet} utilized the weight-sharing SwinTransformer \cite{swintransformer} to construct a backbone for extracting multi-level features, and further enhanced them utilizing the channel attention mechanism. Similarly, Bandara et al. \cite{changeformer} employed Segformer \cite{segformer} to extract multi-level features, which were then subjected to feature differentiation and fed into a decoder to predict detection results. Liu et al. \cite{MSCANet} employed the concept of deep supervision to tokenize visual features of different scales and perform multi-scale supervision. The densely attentive refinement network (DARNet) proposed by Li et al. \cite{li2022DARNet} utilizes a hybrid attention mechanism based on Transformer to model the spatiotemporal relationship of bi-temporal features. Feng et al. \cite{ICIFNet} proposed the intra-scale and inter-scale cross-interaction feature fusion network, which utilizes Transformers to model both intra-scale and inter-scale relationships of bi-temporal features. Building upon this, Feng et al. \cite{DMINet} utilize the concatenated bi-temporal features along the channel as a shared query to model the spatiotemporal relationships between different temporal images. Song et al. \cite{song2022ACABFNet} utilize axial cross-attention based on Transformers to capture global relationships between bi-temporal features. To address the lack of interaction between bi-temporal features during the feature extraction stage, Zhang et al. \cite{zhang2024bifa} proposed a Transformer-based approach for feature extraction in the bi-temporal images.

Although the Transformer-based methods mentioned above have achieved great performance in CD, the complexity of the Transformer in processing images scales quadratically with the length of image patches. This leads to significant computational costs and is not beneficial for tasks like dense prediction, such as CD. In this paper, we integrate Mamba, which is considered an alternative to Transformer due to its linear complexity, into our CD model to mitigate the computational challenges mentioned above.

\subsection{Mamba-based Models in Vision Tasks}

Recently, State Space Models (e.g., Mamba), which exhibit a linear computational complexity for the input sequence length compared to Transformer, have shown potential in effectively modeling long sequences, offering an alternative solution for addressing long-term dependency relationships in visual tasks. Zhu et al. \cite{Vision_mamba} pioneered the application of Mamba in visual tasks. Specifically, to handle position-sensitive image data, the authors proposed Vision Mamba (Vim), which combines position encoding and bidirectional scanning to effectively capture the global context of images. Almost simultaneously, Liu et al. \cite{liu2024vmamba} introduced VMamba, which addresses the position-sensitive challenges by traversing the image space via four-directional scanning (top-left, bottom-right, top-right, and bottom-left). Since then, Mamba-based approaches have mushroomed. The specific structures for processing medical images are proposed based on the Mamba module, including Mamba-Unet \cite{Mamba-Unet}, VM-Unet \cite{Vm-unet}, U-Mamba \cite{U-mamba}, LightM-Unet \cite{LightM-UNet}, SegMamba \cite{Segmamba}. Yang et al. \cite{plainmamba} proposed PlainMamba, which achieves 2D continuous scanning with direction-aware tokens. Pei et al. \cite{pei2024efficientvmamba} improved the scanning method of Mamba by utilizing the technique of dilated convolution, enhancing the efficiency of Mamba. Huang et al. \cite{huang2024localmamba} proposed LocalMamba, which dynamically determines scanning schemes for different layers through a process of dynamic search. Chen et al. \cite{MIM-ISTD} applied the concept of visual sentences and visual words to incorporate Mamba into infrared small target detection. Recently, there have been several methods applying Mamba to remote sensing tasks. Chen et al. \cite{chen2024rsmamba} proposed RSMamba by combining shuffle with forward and backward scanning. Zhao et al. \cite{RS-Mamba} introduce diagonal scanning to process image segmentation and change detection tasks. Around the same time, Chen et al. \cite{chen2024changemamba} utilized multiple scanning methods to learn spatiotemporal relationships between bi-temporal data. 

Although the aforementioned methods have achieved promising results, most of them rely on modifying the scanning methods of Mamba to enhance the global receptive field. However, for dense prediction tasks such as CD, local information plays a crucial role in archiving accurate detection. In this article, we combine the Mamba to propose a simple and efficient structure that integrates both global and local information. Furthermore, we leverage the structure to build a CD model aimed at achieving fine-grained detection.

\begin{figure*}
        \centering
        \includegraphics[width=0.94\textwidth]{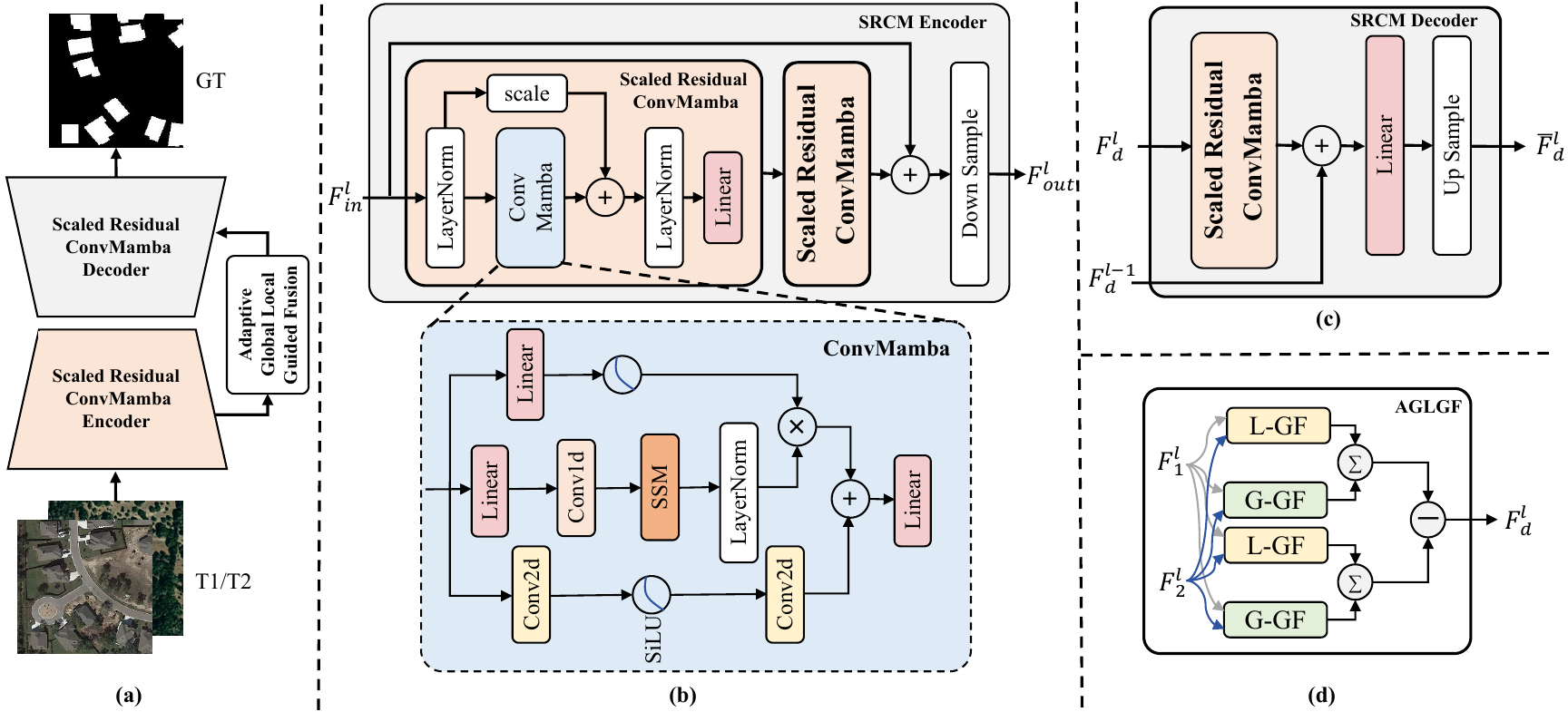}
        \caption{Illustration of our method. (a) is the architecture of the proposed CDMamba. T1 and T2 represent bi-temporal images, and GT means the ground truth. (b) represents the encoder composed of the Scaled Residual ConvMamba (SRCM) block, as well as its main component, the ConvMamba module. $F^l_{in}$, $F^l_{out}$ represent the input and output features of various levels from bi-temporal images. (c) represents the decoder formed by SRCM, where $F^l_{d}$ and $F^{l-1}_{d}$ represent the current level and the previous level differential feature, respectively. And $\overline{F}^l_{d}$ is the feature after multi-level fusion. (d) indicates the Adaptive Global Local Guided Fusion (AGLGF) block, where $F^l_{1}$ and $F^l_{2}$ are bi-temporal features at the same level. $F^l_{d}$ is the differential features at level \emph{l}. The L-GF represents the local-guided feature fusion module and the G-GF means the global-guided feature fusion module. And $\sum$ is the weighted summation,
        }
        \label{fig:CDMamba}
\end{figure*}

\section{CDMamba}
\label{sec:method}

\subsection{Preliminaries}
\label{ssec:Preliminaries}

The recently emerging structured state-space sequence models (SSMs) (e.g., S4) are mostly inspired by linear time-invariant systems. These models map a one-dimensional function or sequence  \(x(t)\in \mathbb{R}\) to \(y(t)\in \mathbb{R}\) through a hidden state \(h(t)\in \mathbb{R}^{N}\). Typically, the system is formulated as a linear ordinary differential equation (ODE):
\begin{equation}
    \label{1}
        {h}^{'}(t) =  \mathbf{A}{h(t)} + \mathbf{B}{x(t)}
\end{equation}
\begin{equation}
    \label{2}
        y(t) =  \mathbf{C}{h(t)}
\end{equation}
where \emph{N} indicates the state size, \(\mathbf{A}\in \mathbb{R}^{N \times N}\), \(\mathbf{B}\in \mathbb{R}^{N \times 1}\) and \(\mathbf{C}\in \mathbb{R}^{1 \times N}\).

Subsequently, to integrate the continuous-time representation into deep learning algorithms, a time-scale parameter \(\mathbf{\Delta}\) is typically introduced to discretize the continuous parameters \(\mathbf{A}\) and \(\mathbf{B}\) using a common zero-order hold (ZOH) approach. The conversion results in discrete parameters \(\bar{\mathbf{A}}\) and \(\bar{\mathbf{B}}\).
\begin{equation}
    \label{3}
        \bar{\mathbf{A}} =  exp(\mathbf{\Delta}\mathbf{A})
\end{equation}
\begin{equation}
    \label{4}
        \bar{\mathbf{B}} =  (\mathbf{\Delta}\mathbf{A})^{-1}(exp(\mathbf{\Delta}\mathbf{A}) - \mathbf{I})(\mathbf{\Delta}\mathbf{B})
\end{equation}

After discretization, Eq. \eqref{1} and Eq. \eqref{2} can be represented as:
\begin{equation}
    \label{5}
        {h}(t) =  \bar{\mathbf{A}}{{h}_{t-1}} + \bar{\mathbf{B}}{{x}_{t}}
\end{equation}
\begin{equation}
    \label{6}
         y(t) =  \mathbf{C}{h_t}
\end{equation}
The final output can be obtained directly through full convolution computation.

However, the parameters of the aforementioned process remain constant for different inputs. To address this limitation, the recently proposed Mamba combines scanning mechanisms with data-dependent learnable parameters \(\mathbf{\Delta}\), \(\bar{\mathbf{B}}\) and \(\mathbf{C}\) to adjust the learned contextual content of the model dynamically. Additionally, the hardware-aware algorithm was proposed to enhance its efficiency on GPUs.

\subsection{Overview}
\label{ssec:Overview}

The architecture of the proposed CDMamba is shown in Fig. \ref{fig:CDMamba}. (a), which is composed of the Scaled Residual ConvMamba Block Encoder, Scaled Residual ConvMamba Decoder and Adaptive Global Local Guided Fusion (AGLGF) block. Given the bi-temporal remote sensing images \(\mathbf{T_1} \in \mathbb{R}^{3 \times H  \times W}\) and \(\mathbf{T_2} \in \mathbb{R}^{3 \times H  \times W}\) where 3 represents the channel dimension, \emph{H} and \emph{W} represent the height and width of the image, respectively. First, \(\mathbf{T_1}\) and \(\mathbf{T_2}\) are fed into the Convolution Stream to respectively extract shallow features, obtaining shallow feature maps \(\mathbf{F_1} \in \mathbb{R}^{C_1 \times H  \times W}\) and \(\mathbf{F_2} \in \mathbb{R}^{C_1 \times H  \times W}\). Subsequently, these features are sent into several cascaded Encoder blocks, whcih consists of Scaled Residual ConvMamba (SRCM), residual connection, and downsampling, to extract bi-temporal features at different scales \(\left\{\mathbf{F_1^i}\right\}_{i=1}^4\) and \(\left\{\mathbf{F_2^i}\right\}_{i=1}^4\). Considering the requirement for the interaction of bi-temporal features in change detection tasks, the obtained multi-scale deep features are individually fed into the AGLGF block, which consists of global/local guided fusion block and adaptive gating, to facilitate learning abundance semantic contexts. Specifically, global/local guided fusion block is utilized to achieve bi-temporal interaction, and adaptive gating is used to perform adaptive fusion. Finally, various scale differential features \(\left\{\mathbf{F_d^i}\right\}_{i=1}^4\) are obtained using the method of absolute subtraction. During the decoding stage, differential features \(\left\{\mathbf{F_d^i}\right\}_{i=1}^4\) are sent into the decoder consisting of SRCM, convolution, and upsampling operations. By fusing with features from adjacent scales, the feature maps are gradually restored to the original image size. Finally, the change detection result is obtained through a linear projection.

\subsection{Scaled Residual ConvMamba Block}
\label{ssec:SRCM}

For dense prediction tasks (e.g., CD), local information plays a crucial role in accurate detection. However, current methods based on Mamba primarily focus on enhancing the model's ability to extract global features by designing different scanning methods, often neglecting the importance of local information. We aim to explore a simple yet effective structure combining Mamba, which integrates both global and local information simultaneously. A straightforward approach is to combine the local features extracted by convolution with the global features extracted by Mamba. Therefore, we proposed the Scaled Residual ConvMamba module, as shown in Fig. \ref{fig:CDMamba}. (b).

Given the input feature \(\mathbf{F_{in}} \in {R}^{L \times C}\), the SRCM module initially applies LayerNorm \cite{Layernormalization} followed by a ConvMamb module to capture global and local spatial features resulting in \(\mathbf{F_{gl}} \in {R}^{L \times C}\). Furthermore, to capture more comprehensive contextual features, the fusion of \(\mathbf{F_{gl}}\) and \(\mathbf{F_{in}}\) is achieved by the scaled residual connection. The fused feature is followed by normalization utilizing LayerNorm and then linear transformation to learn deeper features. The entire process can be described as follows:
\begin{equation}
    \label{7}
         \overset{\sim}{\mathbf{F}} = ConvMamba(LN(\mathbf{F_{in}})) + \alpha\mathbf{F_{in}}
\end{equation}
\begin{equation}
    \label{8}
         \mathbf{F_{out}} = Linear((LN(\bar{\mathbf{F}})))
\end{equation}

Specifically, within the ConvMamba module, there are three branches. The first branch takes the first half of the input feature \(\mathbf{F_{in}}\) along the channel dimension as input \(\mathbf{F_{b1}}\ \in {R}^{L \times \frac{C}{2}}\), and then expands the dimension to \(\mathbf{\lambda{C}}\) through a linear transformation, followed by activation using the SiLU \cite{SiLU} function. The input of the second branch within the ConvMamba module is similar to that of the first branch. It takes the second half of the input feature \(\mathbf{F_{in}}\) along the channel dimension as input \(\mathbf{F_{b2}}\ \in {R}^{L \times \frac{C}{2}}\). Subsequently, the features are sequentially passed through dimension-expanding linear layers, Conv1d layers, SSM, and LayerNorm. Following this, the features extracted from these two branches are fused utilizing the Hadamard product, aiming to capture global features in this manner \cite{mamba}. The input to the third branch is the transformed \(\mathbf{F_{in}}\), where \(\mathbf{F_{in}}\) is reshaped into \(\mathbf{F_{b3}} \in {R}^{C \times H \times W}\). Subsequently, \(\mathbf{F_{b3}}\) is fed into the Conv2d layer followed by activation using the SiLU function, aiming to capture local features. Finally, the local features obtained from the third branch are flattened and fused with the previously captured global features using addition. The fused features are followed by a linear mapping to obtain the feature \(\mathbf{F_{gl}}\) that integrates both global and local information. The overall process is formalized as follows:
\begin{equation}
    \label{9}
         \overset{\sim}{\mathbf{F_1}} = SiLU(Linear(\mathbf{F_{b1}}))
\end{equation}
\begin{equation}
    \label{10}
         \overset{\sim}{\mathbf{F_2}} = LN(SSM(C1(Linear(\mathbf{F_{b2}}))))
\end{equation}
\begin{equation}
    \label{11}
         \overset{\sim}{\mathbf{F_3}} = C2(SiLU(C2(\mathbf{F_{b3}})))
\end{equation}
\begin{equation}
    \label{12}
         \mathbf{F_{gl}} = Linear(\overset{\sim}{\mathbf{F_1}} \odot \overset{\sim}{\mathbf{F_2}} + \overset{\sim}{\mathbf{F_3}})
\end{equation}
where \(\mathbf{\odot}\) represents Hadamard product, C1 and C2 means Conv1d and Conv2d, respectively.

\subsection{Adaptive Global Local Guided Fusion Block}
\label{ssec:AGLGF}

\begin{figure}
        \centering
        \includegraphics[width=0.47\textwidth]{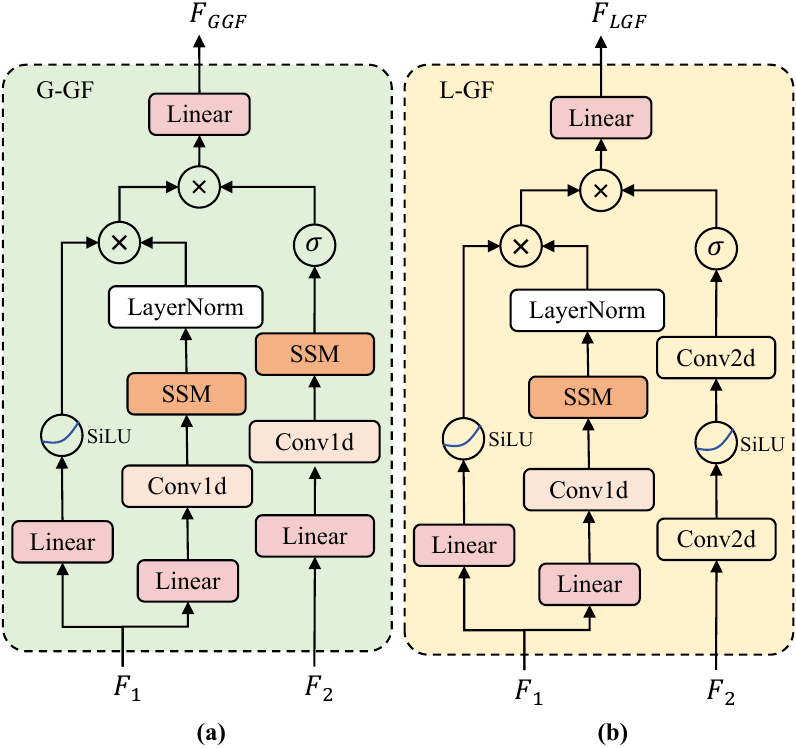}  
        \caption{Illustration of our global-guided feature fusion (G-GF) module and local-guided feature fusion (L-GF) module. $\sigma$ means the gate activation function. $F_1$ and $F_2$ represent bi-temporal features, respectively. And $F_{GGF}$ and $F_{LGF}$ are global-guided and local-guided fused features.}
        \label{fig:AGLGF}
\end{figure}

Considering the requirement for the interaction of bi-temporal features in CD tasks, we propose the Adaptive Global Local Guided Fusion block (AGLGF), as shown in Fig. \ref{fig:CDMamba}. (d), that dynamically combines global-guided and local-guided features to provide more discriminative change features.

Take \(\mathbf{F_{2}}\) guided \(\mathbf{F_{1}}\) for feature fusion as an example. Given the bi-temporal features \(\mathbf{F_{1}}\), \(\mathbf{F_{2}} \in {R}^{L \times C}\) (for simplicity, omitting the stage indices), both are fed into the global-guided feature fusion (G-GF) module, as shown in Fig. \ref{fig:AGLGF}. (a). Specifically, the G-GF module, which is inspired by cross-attention and utilizes a three-branch design, employs the first two branches to process \(\mathbf{F_{1}}\). The first branch utilizes a linear mapping followed by the SiLU activation function. The second branch applies linear mapping, followed by Conv1d and SSM, and then LayerNorm. Finally, the features from the two branches are fused by the Hadamard product to obtain the intermediate feature \(\bar{\mathbf{F_{1}}}\). The third branch is utilized to process the \(\mathbf{F_{2}}\). Initially, \(\mathbf{F_{2}}\) is expanded in dimension through linear transformation. Subsequently, it is sequentially fed into Conv1d layer and SSM to learn global features through scanning \cite{mamba}. Unlike the second branch, an additional gating mechanism is incorporated that controls which features will be activated for guiding \(\mathbf{F_{1}}\). The specific process is formulated as follows:
\begin{equation}
    \label{13}
         \bar{\mathbf{F_{b1}}} = SiLU(Linear(\mathbf{F_{1}})))
\end{equation}
\begin{equation}
    \label{14}
         \bar{\mathbf{F_{b2}}} = LN(SSM(C1(Linear(\mathbf{F_{1}}))
\end{equation}
\begin{equation}
    \label{15}
         \bar{\mathbf{F_{b3}}} = \sigma(SSM(C1(Linear(\mathbf{F_{2}}))
\end{equation}
\begin{equation}
    \label{16}
         \bar{\mathbf{F_1}} = \bar{\mathbf{F_{b1}}} \odot \bar{\mathbf{F_{b2}}}
\end{equation}
\begin{equation}
    \label{17}
         \mathbf{F_{GGF}} = Linear(\bar{\mathbf{F_1}} \odot \bar{\mathbf{F_{b3}}})
\end{equation}

In addition, we propose a local-guided feature fusion (L-GF) module to utilize local features from \(\mathbf{F_{2}}\) for guidance \(\mathbf{F_{1}}\), as shown in Fig. \ref{fig:AGLGF}. (b). Similar to G-GF, L-GF performs the same operation in extracting \(\mathbf{F_1}\) features. However, while extracting \(\mathbf{F_{2}}\) local features in the third branch, \(\mathbf{F_{2}}\) is first transformed as \(\mathbf{F_{2}^{'}} \in \mathbb{R}^{C \times H \times W}\), then fed into the Conv2d layer activated by the activation function. Finally, the feature is flattened and fused with \(\bar{\mathbf{F_1}}\) through the gated mechanism. The specific process is as follows:
\begin{equation}
    \label{18}
         \bar{\mathbf{F_{b3}}} = \sigma(C2(SiLU(C2(\mathbf{{F_{2}^{'}}}))))
\end{equation}
\begin{equation}
    \label{19}
         \mathbf{F_{LGF}} = Linear(\bar{\mathbf{F_1}} \odot \bar{\mathbf{F_{b3}}})
\end{equation}

After obtaining \(\mathbf{F_{GGF}}\) and \(\mathbf{F_{LGF}}\), we employ the dynamic gating mechanism to encourage complementary feature fusion while suppressing redundant features. Specifically, the information from \(\mathbf{F_{GGF}}\) and \(\mathbf{F_{LGF}}\) is compressed into the channel dimension by taking their average. These compressed features are then concatenated along the channel dimension and fed into the linear layer to obtain the values for dynamic gating. Finally, \(\mathbf{F_{GGF}}\) and \(\mathbf{F_{LGF}}\) are fused by weighted sum, and the global and local features of dynamic fusion \(\mathbf{F_{GL}}\) are obtained. The process is as follows:
\begin{equation}
    \label{20}
         \mathbf{F_{concat}} = Concat(mean(\mathbf{F_{GGF}}, \mathbf{F_{LGF}}))
\end{equation}
\begin{equation}
    \label{21}
         \mathbf{G_{score}} = Softmax(Linear(\mathbf{F_concat}))
\end{equation}
\begin{equation}
    \label{22}
         \mathbf{F_{GL}^1} = (\mathbf{G_{score}^{1}} \odot \mathbf{F_{GGF}}) + (\mathbf{G_{score}^{2}} \odot \mathbf{F_{LGF}})
\end{equation}

Similarly, \(\mathbf{F_{GL}^2}\) guided by \(\mathbf{F_1}\) can be obtained by swapping the input positions of \(\mathbf{F_1}\) and \(\mathbf{F_2}\). Finally, \(\mathbf{F_{GL}^1}\) and \(\mathbf{F_{GL}^2}\) are utilized to generate difference features through the absolute subtraction method.

\section{Experimental Results and Analysis} 
\label{sec:experiment}

\subsection{Data description}
\label{ssec:data}

Extensive experiments are conducted on three representative CD datasets to verify the practical performance of the proposed CDMamba.

\subsubsection{Wuhan University}
WHU-CD \cite{ji2018whu} is a dataset tailored for CD tasks, consisting of a pair of 32507×15354 spatial remote sensing images of New Zealand with the resolution of 0.2m/pixels, taken in April 2012 and April 2016, covering an area of 20.5 square kilometers. Since no partitioning strategy was provided in \cite{ji2018whu}, we followed a mainstream approach (e.g., \cite{BIT, zhang2024bifa}) to cut the image into 256 $\times$ 256 patches, and divided them into 6096/762/762 for training/validation/testing.

\subsubsection{Learning, VIsion, and Remote sensing}
LEVIR-CD \cite{Chen2020STA} is a widely utilized CD dataset containing 637 pairs of Google Earth images with the patch size of 1024×1024, the resolution of 0.5m/pixel, and the time-span ranging from 5 to 14 years. The dataset focuses on building-related changes, such as the addition and removal of buildings. We use the method provided by the official source \cite{Chen2020STA} to divide the image into non-overlapping patches of size 256 $\times$ 256 and divide them into 7120/1024/2048 for training/validation/testing.

\subsubsection{LEVIR+-CD}
LEVIR+-CD dataset is an extension of the LEVIR-CD, containing 985 pairs of images with the spatial patch of 1024 $\times$ 1024 pixels. The dataset notably focuses on various types of buildings including urban residential areas, small-scale garages, large warehouses, and more. We cut the images into patches of size 256 $\times$ 256, following the mainstream partitioning method (e.g., \cite{zhang2024bifa}), and divided them into 10192/5568 for training/testing.

\subsection{Experimental setup}
\label{ssec:setup}

\subsubsection{Architecture details}
\label{ssec:architecture}

In the proposed CDMamba, during the encoder stage, the convolutional kernel size of Conv Stream is set to 3 with a stride of 1, and the number of output channels is set to 16 for shallow feature extraction. The number of layers for each four stages in the encoder block ${N_i}$ is set as \(\left\{{1, 2, 2, 4}\right\}\). The spatial resolution of the extracted image features is respectively the same as the original image size, 1/2, 1/4, and 1/8. The channel numbers ${C_i}$ are set as \(\left\{{16, 32, 64, 128}\right\}\). 
In each stage, downsampling is performed using bilinear interpolation. The channel expansion factor $\lambda$ in the ConvMamba module, which includes Linear and Conv2d operations, is set to 2. The channel dimensions in the AGLGF block at different stages are consistent with the ConvMamba at each stage of the encoder. In the decoding stage, the number of decoder blocks at each stage is set to \(\left\{{1, 1, 1}\right\}\). To reduce parameters, the SRCM utilizes depthwise separable convolutions, and upsampling is performed using bilinear interpolation.

\subsubsection{Training details}
\label{ssec:training}

The proposed CDMamba is implemented based on the Pytorch framework and runs on an NVIDIA RTX 4090. For optimization, we utilize the Adam optimizer with an initial learning rate of 1$e$-4, ${\beta_1}$ and ${\beta_2}$ are 0.9, 0.999, respectively. The mini-batch size is set to 6. The total training epochs are 300. The loss function is an addition of the cross-entropy loss and the dice loss \cite{sudre2017dice}.
\begin{equation}
    \label{23}
        L_{total} = \lambda_1{L_{ce}}+\lambda_2{L_{dice}}
\end{equation}
\begin{equation}
    \label{24}
        L_{ce} = -\frac{1}{N}{\sum_{i=1}^N{y_ilog(\hat{y_i})}}
\end{equation}
\begin{equation}
    \label{25}
        L_{dice} = 1-\frac{2\sum_{i=1}^Ny_i\hat{y_i}}{{\sum_{i=1}^Ny_i}+{\sum_{i=1}^N\hat{y_i}}}
\end{equation}
where $\lambda_1$ and $\lambda_2$ denote the coefficients of loss function, ${y_i}$ is the ground truth in the $i$th pixel, $\hat{y_i}$ represents the probability in the $i$th pixel. $N$ indicates the number of pixels.

\subsubsection{Evaluation metrics}
\label{ssec:evaluation}

To evaluate the performance of the proposed CDMamba, we employed five key evaluation metrics, namely overall accuracy (OA), precision (Pre), recall (Rec), F1 score, and intersection over union (IoU). The OA represents the proportion of correctly predicted pixels out of the total pixels. P reflects the proportion of true positive pixels among all pixels predicted as positive. R indicates the proportion of true positive pixels among all truly positive pixels in the ground truth. The F1 Score balances precision and recall by calculating the harmonic mean of P and R. IoU measures the overlap between predicted and ground truth positive regions. The metrics can be individually defined as follows.
\begin{equation}
    \label{26}
        Precision = \frac{TP}{TP+FP}
\end{equation}
\begin{equation}
    \label{27}
        Recall = \frac{TP}{TP+FN}
\end{equation}
\begin{equation}
    \label{28}
        F1 = \frac{2}{Recall^{-1}+Precision^{-1}}
\end{equation}
\begin{equation}
    \label{29}
        IoU = \frac{TP}{TP+FP+FN}
\end{equation}
\begin{equation}
    \label{30}
        OA = \frac{TP+TN}{TP+TN+FP+FN}
\end{equation}
where TP, TN, FP, and FN represent the number of true positives, true negatives, false positives, and false negatives, respectively. It is worth noting that F1 and IoU can better reflect the generalization ability of the model.

\begin{table*}
    \centering
    \caption{Comparison results on the three CD test sets. The top three results are highlighted in \textcolor{red}{red}, \textcolor{green}{green}, \textcolor{blue}{blue}. All results are described in percentage (\%).}
    \resizebox{1\textwidth}{!}{
    \begin{tabular}{c|c|c|c|c}
    \toprule
    \multirow{2}{*}{\textbf{Type}} & 
    \multirow{2}{*}{\textbf{Models}}  &
    \multicolumn{1}{c|}{\textbf{WHU-CD}}  &
    \multicolumn{1}{c|}{\textbf{LEVIR-CD}}  &
    \multicolumn{1}{c}{\textbf{LEVIR+-CD}}  \\
    & & Pre. / Rec. / F1 / IoU / OA & Pre. /  Rec. / F1 / IoU / OA & Pre. /  Rec. / F1 / IoU / OA \\
    \midrule
    \multirow{5}{*}{\textbf{CNN-based}}
    & \makecell[l]{FC-EF$_{18}$ \cite{daudt2018FC-EF}}
    & 92.10 / 90.64 / 91.36 / 84.10 / 99.32  
    & 90.64 / 87.23 / 88.90 / 80.03 / 98.89 
    & 76.49 / 76.32 / 76.41 / 61.82 / 98.08 \\
    
    & \makecell[l]{FC-Siam-Diff$_{18}$ \cite{daudt2018FC-EF}}
    & 87.39 / \textcolor{red}{\textbf{92.36}} / 89.81 / 81.50 / 99.16  
    & 90.81 / 88.59 / 89.69 / 81.31 / 98.96 
    & 80.88 / 77.65 / 79.23 / 65.61 / 98.34\\
     
    & \makecell[l]{FC-Siam-Conc$_{18}$ \cite{daudt2018FC-EF}}
    & 86.57 / 91.11 / 88.78 / 79.83 / 99.08  
    & 91.41 / 88.43 / 89.89 / 81.64 / 98.98 
    & \textcolor{blue}{\textbf{81.12}} / 77.16 / 79.09 / 65.42 / 98.33 \\
     
    & \makecell[l]{IFNet$_{20}$ \cite{zhang2020DSIFN}}
    & 91.51 /  88.01 / 89.73 / 81.37 / 99.20  
    & 89.62 /  86.65 / 88.11 / 78.75 / 98.81 
    & \textcolor{green}{\textbf{81.79}} /  78.40 / 80.06 / 66.76 / 98.41\\
     
    & \makecell[l]{SNUNet$_{21}$ \cite{fang2021snunet}}
    & 84.70 / 89.73 / 87.14 / 77.22 / 98.95 
    & 89.73 / 87.47 / 88.59 / 79.51 / 98.85 
    & 78.90 / 78.23 / 78.56 / 64.70 / 98.26 \\
    \midrule
    \multirow{7}{*}{\textbf{Transformer-based}}
    & \makecell[l]{SwinUnet$_{22}$ \cite{zhang2022swinsunet}}
    & 92.44 / 87.56 / 89.93 / 81.71 / 99.22 
    & 89.11 / 86.47 / 87.77 / 78.21 / 98.77 
    & 77.65 / 78.98 / 78.31 / 64.35 / 98.22\\
     
    & \makecell[l]{BIT$_{22}$ \cite{BIT}}
    & 91.84 / \textcolor{green}{\textbf{91.95}} / 91.90 / 85.01 / 99.35 
    & \textcolor{green}{\textbf{92.07}} / 88.08 / 90.03 / 81.87 / \textcolor{blue}{\textbf{99.01}} 
    & 80.50 / 81.41 / \textcolor{blue}{\textbf{80.95}} / \textcolor{blue}{\textbf{68.00}} / \textcolor{blue}{\textbf{98.43}}\\
    
    & \makecell[l]{ChangeFormer$_{22}$ \cite{changeformer}}
    & 93.73 / 87.11 / 90.30 / 82.32 / 99.26
    & 90.68 / 87.04 / 88.83 / 79.90 / 98.88
    & 77.32 / 77.75 / 77.54 / 63.31 / 98.16\\
      
    & \makecell[l]{MSCANet$_{22}$ \cite{MSCANet}}
    & 93.47 / 89.16 / 91.27 / 83.94 / 99.32 
    & 90.02 / 88.71 / 89.36 / 80.77 / 98.92 
    & 76.92 / \textcolor{red}{\textbf{83.69}} / 80.16 / 66.89 / 98.31\\
      
    & \makecell[l]{Paformer$_{22}$ \cite{liu2022Paformer}}
    & \textcolor{blue}{\textbf{94.28}} / 90.38 / 92.29 / 85.69 / 99.40 
    & 91.34 / 88.07 / 89.68 / 81.29 / 98.96 
    & 79.89 / \textcolor{green}{\textbf{82.96}} / \textcolor{green}{\textbf{81.40}} / \textcolor{green}{\textbf{68.63}} / \textcolor{green}{\textbf{98.45}} \\
      
    & \makecell[l]{DARNet$_{22}$ \cite{li2022DARNet}}
    & 91.99 / 91.17 / 91.58 / 84.46 / 99.33 
    & \textcolor{red}{\textbf{92.19}} / \textcolor{blue}{\textbf{88.99}} / \textcolor{green}{\textbf{90.56}} / \textcolor{green}{\textbf{82.76}} / \textcolor{green}{\textbf{99.05}} 
    & 77.84 / 78.42 / 78.13 / 64.11 / 98.21 \\
      
    & \makecell[l]{ACABFNet$_{23}$ \cite{song2022ACABFNet}}
    & 91.57 / 90.86 / 91.21 / 83.84 / 99.31 
    & 90.11 / 88.27 / 89.18 / 80.48 / 98.91 
    & 72.85 / 80.91 / 76.67 / 62.17 / 97.99 \\
    \midrule
    \multirow{3}{*}{\textbf{Mamba-based}}  
    & \makecell[l]{RS-Mamba$_{24}$ \cite{RS-Mamba}}
    & \textcolor{green}{\textbf{95.50}} / 90.24 / \textcolor{green}{\textbf{92.79}} / \textcolor{green}{\textbf{86.55}} / \textcolor{green}{\textbf{99.44}} 
    & 91.36 / 88.23 / 89.77 / 81.44 / 98.97 
    & 79.67 / \textcolor{blue}{\textbf{82.19}} / 80.91 / 67.95 / 98.42 \\
    
    & \makecell[l]{ChangeMamba$_{24}$ \cite{chen2024changemamba}}
    & 94.21 / 90.94 / \textcolor{blue}{\textbf{92.55}} / \textcolor{blue}{\textbf{86.13}} / \textcolor{blue}{\textbf{99.42}} 
    & \textcolor{blue}{\textbf{91.59}} / \textcolor{green}{\textbf{88.78}} / \textcolor{blue}{\textbf{90.16}} / \textcolor{blue}{\textbf{82.09}} / 99.01 
    & 79.64 / 81.92 / 80.77 / 67.74 / 98.41 \\
    
    & \makecell[l]{CDMamba}
    & \textcolor{red}{\textbf{95.58}} / \textcolor{blue}{\textbf{92.01}} / \textcolor{red}{\textbf{93.76}} / \textcolor{red}{\textbf{88.26}} / \textcolor{red}{\textbf{99.51}} 
    & 91.43 / \textcolor{red}{\textbf{90.08}} / \textcolor{red}{\textbf{90.75}} / \textcolor{red}{\textbf{83.07}} / \textcolor{red}{\textbf{99.06}} 
    & \textcolor{red}{\textbf{85.11}} / 81.00 / \textcolor{red}{\textbf{83.01}} / \textcolor{red}{\textbf{70.95}} / \textcolor{red}{\textbf{98.65}}\\
   \bottomrule
    \end{tabular}
    }
    \label{tab:comparison_sotas1}
\end{table*}

\subsection{Performance comparison}
\label{ssec:performance}

To verify the effectiveness of CDMamba in CD tasks, some state-of-the-art methods are selected for comparison in this section, including CNN-based methods (FC-EF \cite{daudt2018FC-EF}, FC-Siam-Diff \cite{daudt2018FC-EF}, FC-Siam-Conc \cite{daudt2018FC-EF}, IFNet \cite{zhang2020DSIFN} and SNUNet \cite{fang2021snunet}), Transformer-based methods (SwinUnet \cite{zhang2022swinsunet}, Changeformer \cite{changeformer}, BIT \cite{BIT}, MSCANet \cite{MSCANet}, Paformer \cite{liu2022Paformer}, DARNet \cite{li2022DARNet}, ACABFNet \cite{song2022ACABFNet}, and DMINet \cite{DMINet}). And the Mamba-based methods (RS-Mamba \cite{RS-Mamba}, ChangeMamba \cite{chen2024changemamba}).

For a fair comparison, all methods are trained under the same conditions based on the officially published Pytorch code.

\subsubsection{Quantitative results}
\label{ssec:quantitative}
In numerical terms, Table \ref{tab:comparison_sotas1} presents the overall performance of all comparative methods on the WHU-CD, LEVIR-CD, LEVIR+-CD test sets. The red font represents the best result, followed by green, and blue indicates the third-best result. Evidently, whether compared to CNN-based methods, Transformer-based methods, or the latest Mamba-based methods, our proposed CDMamba demonstrates superior performance. Specifically, compared to CNN-based methods, on the WHU-CD dataset, although CDMamba exhibits a relatively lower Rec. metric compared to FC-Saim-Diff, it outperforms FC-Saim-Diff on other metrics, indicating that CDMamba possesses superior accuracy in detecting changed regions. In contrast to Transformer-based methods, although CDMamba performs relatively lower Pre. metric than DARNet, it is superior to DARNet in other measures, on the LEVIR-CD dataset. This demonstrates that CDMamba is more comprehensive in detecting areas of change. In comparison with the recently proposed Mamba-based methods (RS-Mamba and ChangeMamba), the CDMamba proposed in this article, although not optimal in terms of some precision and recall measures, achieves the best performance in terms of F1 score and IoU. For the F1 score, there are improvements of 0.97\%/1.21\%, 0.98\%/0.59\%, and 2.10\%/2.24\% on the WHU-CD, Levir-CD, and Levir+-CD datasets, respectively. This indicates that CDMamba provides a more balanced performance in detecting changed regions. In summary, the above quantitative analysis proves that effective combination of local and global information is essential for dense prediction tasks such as CD.

\subsubsection{Qualitative results}
\label{ssec:qualitative}

\begin{figure*}
    \centering
    \includegraphics[width=0.98\textwidth]{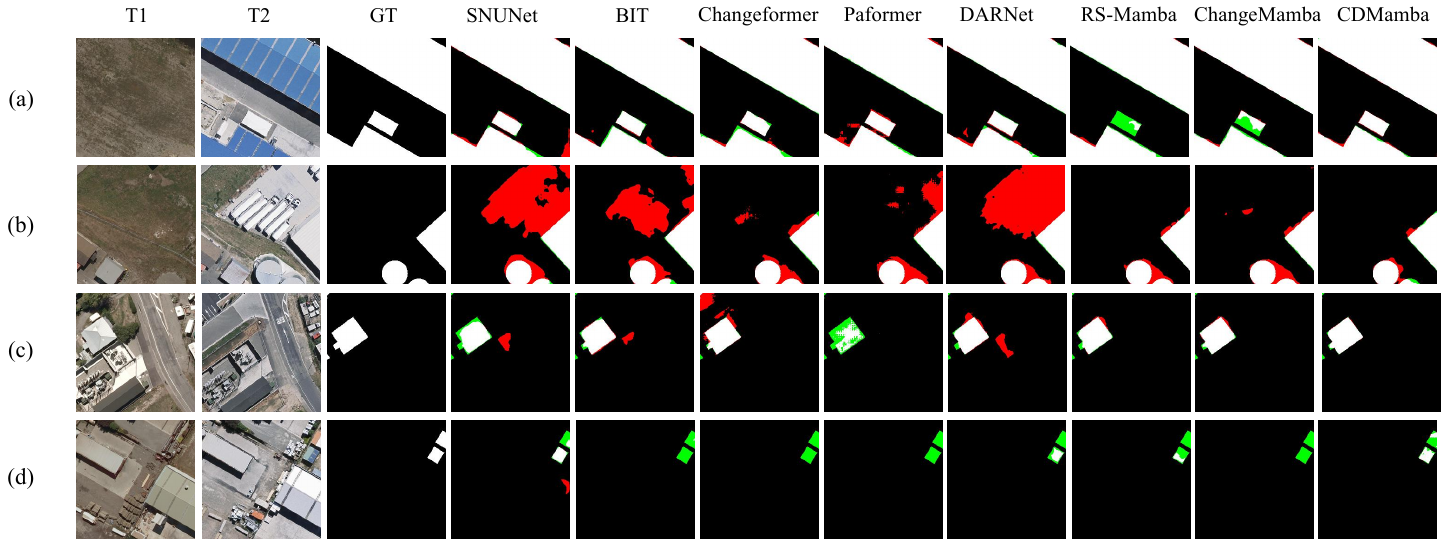}
    \caption{Visualization results of different methods on the WHU-CD test set. (a)-(d) are representative samples. White represents a true positive, black is a true negative, {\color{red} red} indicates a false positive, and {\color{green} green} stands as a false negative.}
    \label{fig:whu}
\end{figure*}

\begin{figure*}
    \centering
    \includegraphics[width=0.98\textwidth]{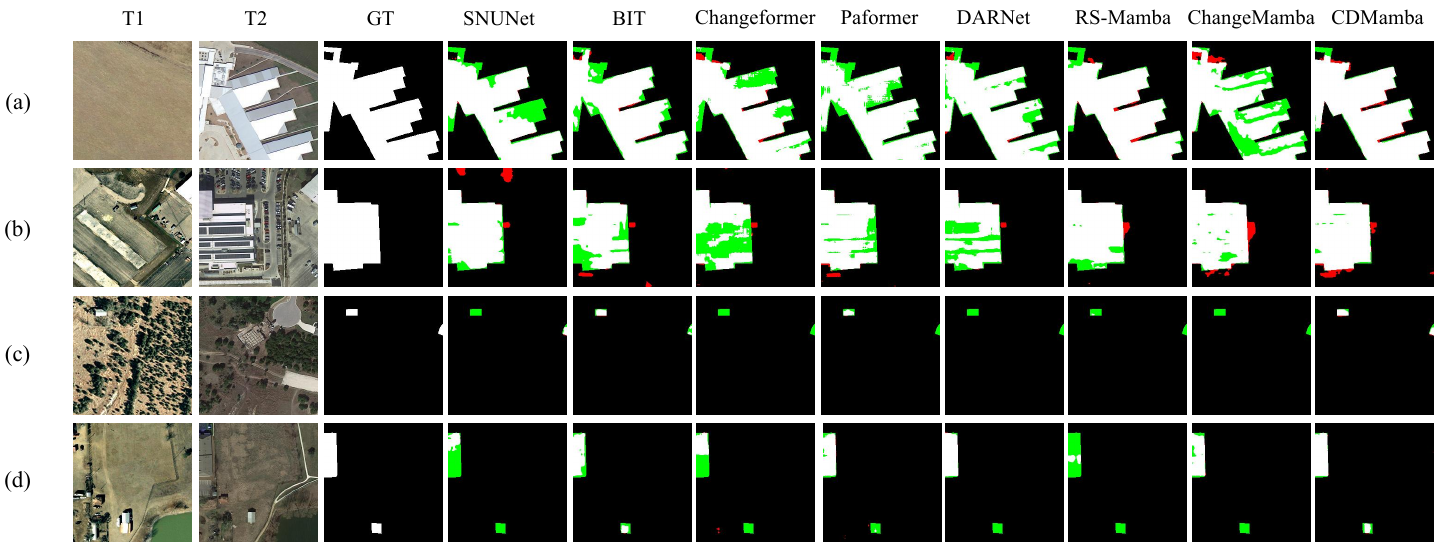}
    \caption{Visualization results of different methods on the LEVIR-CD test set. (a)-(d) are representative samples. White represents a true positive, black is a true negative, {\color{red} red} indicates a false positive, and {\color{green} green} stands as a false negative.}
    \label{fig:levir}
\end{figure*}

\begin{figure*}
    \centering
    \includegraphics[width=0.98\textwidth]{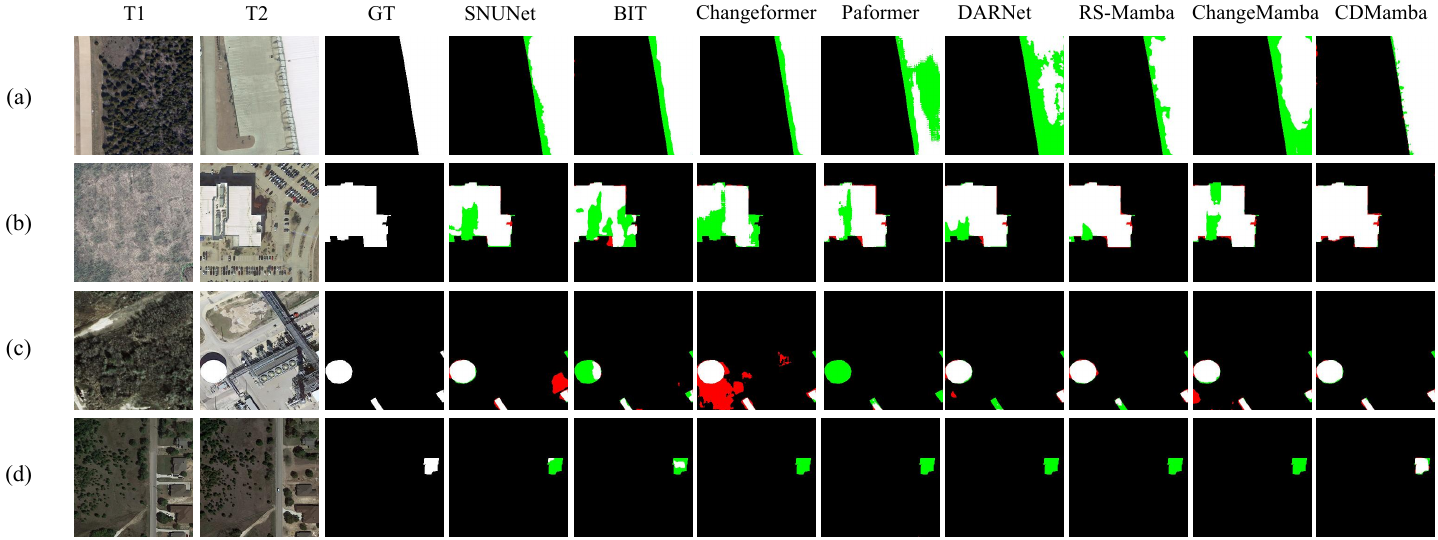}
    \caption{Visualization results of different methods on the LEVIR+-CD test set. (a)-(d) are representative samples. White represents a true positive, black is a true negative, {\color{red} red} indicates a false positive, and {\color{green} green} stands as a false negative.}
    \label{fig:levir+}
\end{figure*}

To further illustrate the validity of our proposed method, qualitative analyses are conducted on WHU-CD, LEVIR-CD, LEVIR+-CD test sets (Fig. \ref{fig:whu}-\ref{fig:levir+}), where distinct colors are assigned to identify the correctness or incorrectness of the detection, including TP (white), TN (black), FP (red) and FN (green).

Visualization on WHU-CD (Fig. \ref{fig:whu}): Several representative samples are selected for visualization comparison. For instance, Fig. \ref{fig:whu}(a) and Fig. \ref{fig:whu}(b) depict scenarios with large-scale changes in buildings, while Fig. \ref{fig:whu}(c) and Fig. \ref{fig:whu}(d) illustrate the variations of small buildings in complex scenes. From Fig. \ref{fig:whu}(a), it is evident that our CDMamba outperforms other competitors. Compared with CNN-based and Transformer-based methods, which have obvious missed and false detections at the change edge, our method has a more detailed edge structure. In contrast to Mamba-based methods that show missed detections in small change areas, our CDMamba provides more accurate detection results. As shown in Fig. \ref{fig:whu}(b), compared to methods based on CNN and Transformer that are more susceptible to interference from irrelevant changes, Mamba-based methods achieve more robust results. However, despite RS-Mamba and ChangeMamba achieving relatively satisfactory results, they suffer from severe false detections. In contrast, our CDMamba achieves more comprehensive detection results. Comparing the detection results of various methods in Fig. \ref{fig:whu}(d), while most models completely miss the change areas, our CDMamba can detect small change areas even in complex scenes. In summary, our CDMamba, leveraging the advantages of both global and local modeling, demonstrates a stronger ability to resist interference from irrelevant changes and provides more refined local detection capabilities.

Visualization on LEVIR-CD (Fig. \ref{fig:levir}): We adopted a similar approach by selecting several representative samples for comparison on the LEVIR-CD dataset. Fig. \ref{fig:levir}(a) and Fig. \ref{fig:levir}(b) display the changing scenarios of large-scale buildings, and Fig. \ref{fig:levir}(c) and Fig. \ref{fig:levir}(d) indicate scenes of small-scale building changes. As shown in Fig. \ref{fig:levir}(a) and Fig. \ref{fig:levir}(b), when detecting large irregular buildings, our CDMamba outperforms models based on CNN, Transformer, or Mamba methods. Furthermore, CDMamba also achieves excellent results in the scenarios of small-scale building variations, as shown in Fig. \ref{fig:levir}(c) and Fig. \ref{fig:levir}(d). Compared to RS-Mamba and ChangeMamba, which miss small change areas, CDMamba can detect these regions more effectively. This phenomenon may be due to the integration of local feature extraction capabilities, which makes it more sensitive to small changes.

Visualization on LEVIR+-CD (Fig. \ref{fig:levir+}): In the LEVIR+-CD, we also selected several representative samples. Fig. \ref{fig:levir+}(a) and Fig. \ref{fig:levir+}(b) represent the change scenes of large buildings, while Fig. \ref{fig:levir+}(c) and Fig. \ref{fig:levir+}(d) show small change area scenes. It is evident that our CDMamba achieves the best results in all scenarios. Notably, in the small change area scenario shown in Fig. \ref{fig:levir+}(d), where methods based on CNN, Transformer, and Mamba almost completely miss the changes, the detection results of our CDMamba are nearly identical to the ground truth. This further demonstrates the effectiveness of CDMamba.

\begin{table}
    \centering
    \caption{Comparison results on model efficiency. We report the number of parameters (Params.) and the training time (Time) for a single epoch on LEVIR+-CD.}
    \begin{tabular}{c|c|cc}
    \toprule
    Type & Model & Params. (M) & Time (Min)\\
    \midrule
        \multirow{5}{*}{CNN-based} 
        & \makecell[l]{FC-EF$_{18}$} & 1.35 & 1.61\\
        & \makecell[l]{FC-Siam-Diff$_{18}$} & 1.34  & 1.50\\
        & \makecell[l]{FC-Siam-Conc$_{18}$} & 1.54 & 1.77\\
        & \makecell[l]{IFNet$_{20}$} & 50.71 & 5.02\\
        & \makecell[l]{SNUNet$_{21}$} & 1.35 & 1.65\\
    \midrule
        \multirow{7}{*}{Transformer-based}
        & \makecell[l]{SwinUnet$_{22}$} & 30.28 & 3.01\\
        & \makecell[l]{BIT$_{22}$} & 3.04 & 2.8\\
        & \makecell[l]{Changeformer$_{22}$} & 41.02 & 20.45\\
        & \makecell[l]{MSCANet$_{22}$} & 16.42 & 6.28\\
        & \makecell[l]{Paformer$_{22}$} & 16.13 & 2.12\\
        & \makecell[l]{DARNet$_{22}$} & 15.09 & 12.53\\
        & \makecell[l]{ACABFNet$_{23}$} & 102.32 & 5.12\\
    \midrule
        \multirow{3}{*}{Mamba-based}
        & \makecell[l]{RS-Mamba$_{24}$} & 51.95 & 8.47\\
        & \makecell[l]{ChangeMamba$_{24}$} & 48.57 & 11.66\\
        & \makecell[l]{CDMamba}  & 11.90 & 13.58\\
    \bottomrule
    \end{tabular}
    \label{tab:model_efficiency}
\end{table}

\subsubsection{Model efficiency}
\label{ssec:efficiency}

To further validate the efficiency of the proposed model, Table \ref{tab:model_efficiency} presents the model parameters (Params.) and the time taken to train one epoch on the LEVIR+-CD dataset (Time). Compared to the Transformer-based method Changeformer, our CDMamba outperforms in both parameters and training time, demonstrating higher efficiency. Additionally, compared to the Mamba-based methods RS-Mamba and ChangeMamba, our CDMamba has a lightweight structure, though its training time is slightly higher than theirs. This is because when RS-Mamba and ChangeMamba process the input image for 4$\times$ downsampling before further processing, while CDMamba directly processes the original-sized image before subsequent operations, resulting in a slightly longer training time compared to the former two methods.

\begin{figure*}
    \centering
    \includegraphics[width=0.7\textwidth]{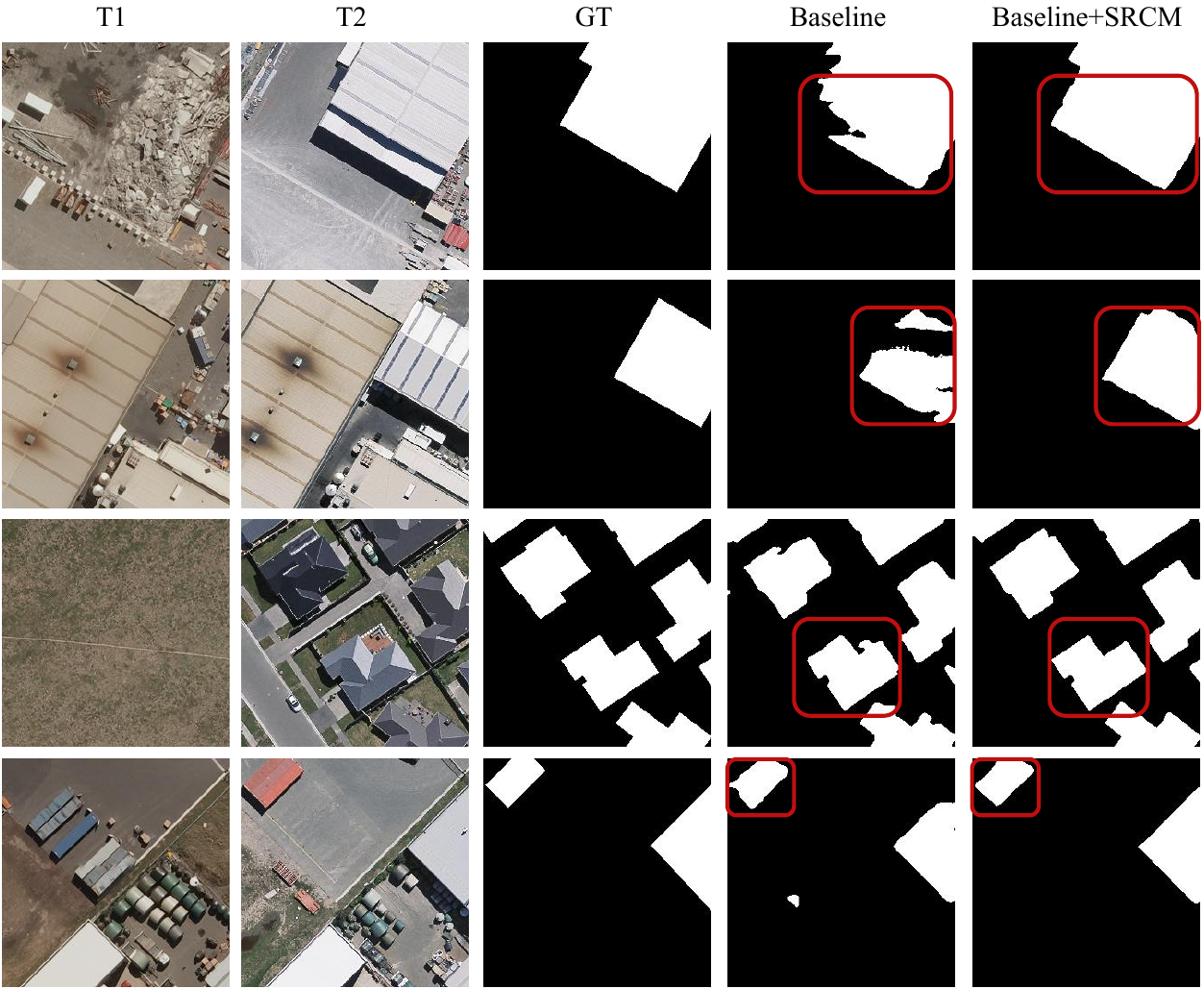}
    \caption{Visualization results of prediction results on the WHU-CD test set. Baseline represents the prediction results of the baseline model and Baseline+SRCM is the prediction results with the addition of SRCM}
    \label{fig:baseline_srcm}
\end{figure*}

\begin{figure*}
    \centering
    \includegraphics[width=0.9\textwidth]{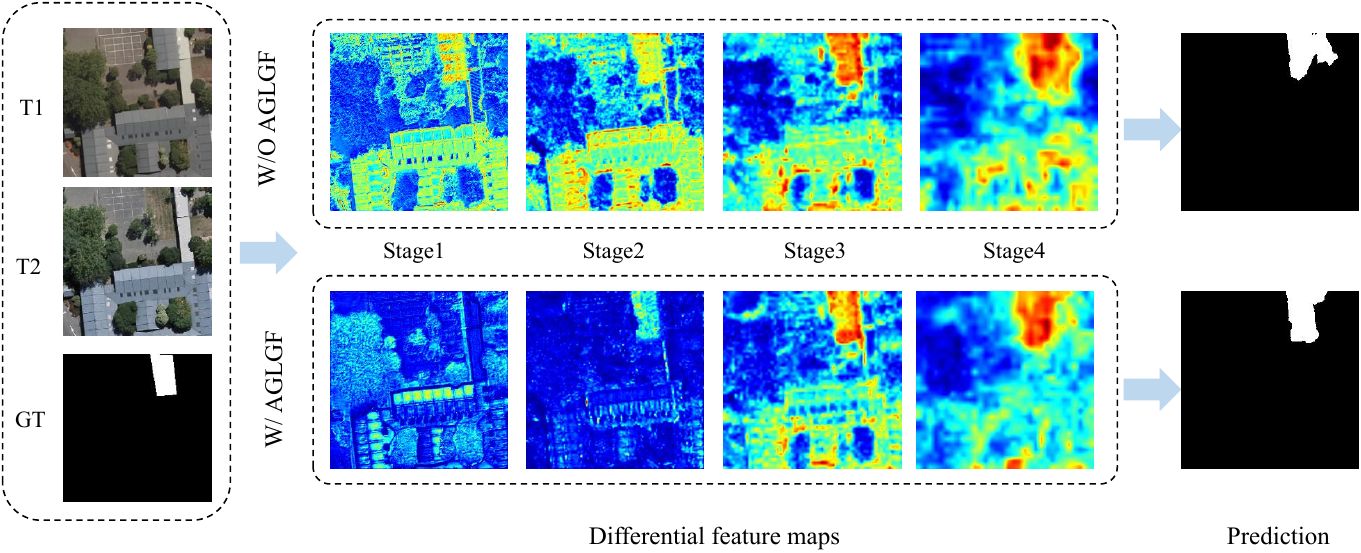}
    \caption{Visualization results of the differential feature maps on the WHU-CD test set. W/O AGLGF denotes the CDMamba without the AGLGF module, and W/ AGLGF represents CDMamba. \textcolor{red}{Red} denotes higher attention values, and \textcolor{blue}{blue} denotes lower values.}
    \label{fig:stagevis}
\end{figure*}

\subsection{Ablation studies}
\label{ssec:ablation}

In this section, we conducted a series of experiments on the WHU-CD dataset to investigate the impact of each component and parameter setting in our proposed method on model performance, as shown in Table \ref{tab:ablation_module}-\ref{tab:ablation_loss}.

\subsubsection{Effects of Different Components in CDMamba}
\label{ssec: Different Components}

To validate the effectiveness of the key modules in CDMamba, we designed eight ablation experiments. Additionally, the original Mamba framework is configured to match the structure of CDMamba, serving as the baseline for comparison. As shown in Table \ref{tab:ablation_module}, the experimental results are superior to the baseline, regardless of whether the key modules are added individually or in combination with each other. The key metrics for change detection, F1 and IoU, have improved by 6.45\% and 10.78\%, respectively. This substantial enhancement demonstrates the importance of effectively integrating global and local features, as well as adaptive differential feature fusion, for change detection tasks. It is noteworthy that after adding the SRCM module, the detection performance of the model improved significantly. We believe this phenomenon quantitatively demonstrates the crucial role of global-local information fusion in accurately detecting tasks of dense prediction, such as change detection. To further validate this intuition, we visualized the results of the Baseline and Baseline+SRCM, as shown in Fig. \ref{fig:baseline_srcm}. It can be observed that regardless of complex scenes with other building interferences or in scenarios with lots of added-in buildings, the results of Baseline+SRCM outcomes with clearer structures and edges. The above phenomenon qualitatively demonstrates the crucial role of integrating global and local information for dense prediction tasks (e.g., CD).

\begin{table}
    \centering
    \caption{Ablation study on Different Components.}
    \resizebox{0.45\textwidth}{!}{
    \begin{tabular}{l|c}
  \toprule
    \multicolumn{1}{c|}{} &
    \multicolumn{1}{c}{\textbf{WHU-CD}}\\
    Model & Pre. / Rec. / F1 / IoU / OA \\
    \midrule
    Baseline
    & 92.36 / 82.78 / 87.31 / 77.48 / 99.04 \\
    \midrule
    +SRCM 
    & 94.97 / 90.77 / 92.83 / 86.61 / 99.44 \\
    +G-GF 
    & 93.02 /  86.80 / 89.81 / 81.50 / 99.21 \\
    +L-GF 
    & 93.23 / 89.12 / 91.14 / 83.72 / 99.31 \\
    +AGLGF 
    & 93.33 / 89.40 / 91.32 / 84.03 / 99.32 \\
    +SRCM+G-GF 
    & 95.19 / 91.69 / 93.41 / 87.63 / 99.48 \\
    +SRCM+L-GF 
    & 95.14 / 91.85 / 93.46 / 87.73 / 99.49 \\
    \midrule
    CDMamba 
    & 95.58 / 92.01 / 93.76 / 88.26 / 99.51 \\
   \bottomrule
    \end{tabular}
    }
    \label{tab:ablation_module}
\end{table}

\begin{table*}
    \centering
    \caption{Ablation study on different stages of AGLGF. And S stands for the stage.}
    \resizebox{0.7\textwidth}{!}{
    \begin{tabular}{l|c|c|c|c|c}
  \toprule
    \multicolumn{1}{c|}{} &
    \multicolumn{1}{c|}{} &
    \multicolumn{1}{c|}{} &
    \multicolumn{1}{c|}{} &
    \multicolumn{1}{c|}{} &
    \multicolumn{1}{c}{\textbf{WHU-CD}} \\
    Model & S1 & S2 & S3 & S4 & Pre. / Rec. / F1 / IoU / OA \\
    \midrule
    CDMamba w/o AGLGF
    & {$\times$}
    & {$\times$}
    & {$\times$}
    & {$\times$}
    & 94.97 / 90.77 / 92.83 / 86.61 / 99.44 \\
    \midrule
    CDMamba 
    & $\checkmark$
    & {$\times$}
    & {$\times$}
    & {$\times$}
    & 93.34 / 93.09 / 93.22 / 87.30 / 99.46 \\
    CDMamba 
    & $\checkmark$
    & $\checkmark$
    & {$\times$}
    & {$\times$}
    & 95.58 / 92.01 / 93.76 / 88.26 / 99.51 \\
    CDMamba 
    & $\checkmark$
    & $\checkmark$
    & $\checkmark$
    & {$\times$}
    & 94.97 / 92.44 / 93.69 / 88.13 / 99.50 \\
    CDMamba 
    & $\checkmark$
    & $\checkmark$
    & $\checkmark$
    & $\checkmark$
    & 95.49 / 91.71 / 93.56 / 87.91 / 99.50 \\
   \bottomrule
    \end{tabular}
    }
    \label{tab:abalation_AGLGF}
\end{table*}

\subsubsection{Effects of different stages of AGLGF}
\label{ssec: different stages of AGLGF}

To further explore the impact of AGLGF at different stages on detection performance, we conducted experiments as shown in Table \ref{tab:abalation_AGLGF}. W/O AGLGF denotes the CDMamba without the AGLGF module, while S1-S4 represents the stages where the AGLGF module is added. With the gradual addition of AGLGF, the performance of the model (F1/IoU) continues to improve until the second stage reaches its best performance. However, further addition of the AGLGF leads to a decline in the performance of the model. This could be attributed to the difficulty of deep semantic features in providing detailed information, making it challenging for the model to learn differential features through guidance, thereby affecting the performance of the model. To provide a more intuitive explanation, we visualized the differential features between W/O AGLGF and W/ AGLGF at the four stages on the WHU-CD test set. As shown in Fig. \ref{fig:stagevis}, adding AGLGF at shallower stages produces better visualization results. Specifically, adding AGLGF at shallower stages allows the model to pay more attention to the changed regions, (for example, in the visualization of Stage 2, W/O AGLGF mostly focuses on the entire image, while W/AGLGF focuses on the changed areas) which can provide better guidance for subsequent stages.

\begin{table}
    \centering
    \caption{Ablation study on different gate activation.}
    \resizebox{0.45\textwidth}{!}{
    \begin{tabular}{l|c}
  \toprule
    \multicolumn{1}{c|}{} &
    \multicolumn{1}{c}{\textbf{WHU-CD}} \\
    Activate Gate & Pre. / Rec. / F1 / IoU / OA \\
    \midrule
    SiLU
    & 95.45 / 91.87 / 93.63 / 88.02 / 99.50 \\
    ReLU 
    & 95.58 / 92.01 / 93.76 / 88.26 / 99.51 \\
    LekyRelu 
    & 95.00 / 91.47 / 93.20 / 87.27 / 99.47 \\
    Sigmoid 
    & 95.17 / 91.77 / 93.45 / 87.69 / 99.48 \\
   \bottomrule
    \end{tabular}
    }
    \label{tab:ablation_gate}
\end{table}

\begin{table}
    \centering
    \caption{Ablation study on different dimensions.}
    \resizebox{0.45\textwidth}{!}{
    \begin{tabular}{l|c}
  \toprule
    \multicolumn{1}{c}{} &
    \multicolumn{1}{c}{\textbf{WHU-CD}} \\
    Dims & Pre. / Rec. / F1 / IoU / OA \\
    \midrule
    d-model
    & 94.86 / 91.67 / 93.24 / 87.33 / 99.47 \\
    1.5$\times$d-model
    & 94.68 / 92.58 / 93.63 / 88.02 / 99.50 \\
    2$\times$d-model
    & 95.58 / 92.01 / 93.76 / 88.26 / 99.51 \\
   \bottomrule
    \end{tabular}
    }
    \label{tab:ablation_dims}
\end{table}

\begin{table}
    \centering
    \caption{Ablation study on different coefficients of loss function}
    \resizebox{0.43\textwidth}{!}{
    \begin{tabular}{c|c|c}
    \toprule
    \multicolumn{1}{c|}{} &
    \multicolumn{1}{c|}{} &
    \multicolumn{1}{c}{\textbf{WHU-CD}}\\
    $\lambda_1$ & $\lambda_2$ & Pre. / Rec. / F1 / IoU / OA \\
        
    \midrule
          1
          & 0
          & 95.47 / 91.84 / 93.62 / 88.01 / 96.68 \\
          0
          & 1
          & 91.06 / 92.35 / 91.70 / 84.68 / 99.33 \\
          0.5
          & 0.5
          & 95.58 / 92.01 / 93.76 / 88.26 / 99.51 \\
          0.5
          & 1
          & 95.04 / 91.94 / 93.46 / 87.73 / 99.49 \\
          1
          & 0.5
          & 94.96 / 91.50 / 93.20 / 87.27 / 99.47 \\
    \bottomrule
    \end{tabular}
    }
    \label{tab:ablation_loss}
\end{table}

\subsubsection{Effects of Different Gate Activation}
\label{Effects of Different Gate Activation}

To further explore the influence of different gate activation methods on model performance, we conducted experiments as shown in Table \ref{tab:ablation_gate}. Replacing $\sigma$ in G-GF and L-GF with different activation functions. It can be observed that non-saturating gate activation functions tend to achieve better results compared to saturating gate activation functions. This could be owing to saturating activation functions that tend to lose detailed features of the input, making it difficult for the model to achieve efficient guidance. Simultaneously, we found that achieving the best performance only requires the relatively simple ReLU gate activation. Therefore, ReLU is chosen as the final gate activation for the model.

\subsubsection{Effects of Different Dimensions}
\label{Effects of Different Dimensions}

To explore the impact of convolutions with different dimensions in L-GF on model performance, we conducted experiments as shown in Table \ref{tab:ablation_dims}. Where d-model represents the feature dimensions in different stages (for detailed settings, refer to \ref{ssec:architecture}). 1.5$\times$d-model indicates expanding the feature dimensions to 1.5 times their original size. Similarly, 2$\times$d-model means doubling the feature dimensions of the model. With the continuous expansion of the feature dimension, the performance of the model also gradually improves. However, when expanding the feature dimensions from 1.5$\times$d-model to 2$\times$d-model, the rate of performance improvement begins to slow down. Therefore, we select the 2$\times$d-model, which achieves the best results, as the final dimension for the module.

\subsubsection{Coefficients of Loss Function} 

To validate the impact of different loss function coefficients on model performance, we conducted corresponding experiments on the WHU-CD dataset. The experimental results are shown in Table \ref{tab:ablation_loss}, where $\lambda_1$ represents the coefficients of cross-entropy loss and $\lambda_2$ indicates the coefficients of dice loss. The experimental results show that the model achieves the best performance when $\lambda_1$ and $\lambda_2$ are relatively balanced. Therefore, we choose $\lambda_1$=$\lambda_2$=0.5 as the final loss function coefficients.

\section{Conclusion}
\label{sec:conclusion}

In this paper, we propose a new model called CDMamba, which effectively combines global and local features for addressing CD tasks. Specifically, to address the challenge that the current Mamba-based method lacks detailed features and struggles to achieve precise detection in dense prediction tasks (e.g., CD), we propose the Scaled Residual ConvMamba (SRCM) block, which combines the ability of Mamba to extract global features with the capability of convolution to extract local clues, for capturing more comprehensive image features. Furthermore, considering the requirement for bi-temporal feature interaction in CD, an Adaptive Global Local Guided Fusion (AGLGF) block is designed to facilitate the interaction of bi-temporal features guided by global and local features. In our intuition, guiding the feature fusion utilizing features from the other temporal can further acquire more discriminative difference features. Many ablation experiments have verified the effectiveness of each module. Meanwhile, experimental results on three public datasets (WHU-CD, LEVIR-CD and LEVIR+CD) show that our method is advantageous over other state-of-the-art methods. In future work, we will explore the Mamba architecture for dense prediction tasks in remote sensing images by incorporating self-supervised learning methods.





\ifCLASSOPTIONcaptionsoff
  \newpage
\fi

{\small
\bibliographystyle{IEEEtran}
\bibliography{refs}
}


\end{document}